\documentclass[10pt,twocolumn,letterpaper]{article}

\usepackage[pagenumbers]{cvpr}

\definecolor{cvprblue}{rgb}{0.21,0.49,0.74}
\usepackage[pagebackref,breaklinks,colorlinks,allcolors=cvprblue]{hyperref}
\usepackage{bbm}
\usepackage{booktabs}
\usepackage{makecell}
\usepackage{tabularx}
\usepackage{pifont}
\usepackage{multirow}
\usepackage[table]{xcolor}
\renewcommand{\arraystretch}{1.1}
\definecolor{rowgray}{gray}{0.90}

\newcommand{\xmark}{\textcolor{red!80!black}{\ding{56}}}
\newcommand{\cmark}{\textcolor{green!70!black}{\ding{52}}}

\def\paperID{}
\def\confName{CVPR}
\def\confYear{2026}

\title{How Much Future Helps? A Controlled Study of Future-Privileged Supervision for Causal Egocentric Gaze Estimation}

\author{
Jia Li$^{1}$, Wenjie Zhao$^{1}$, Fnu Atisri$^{1}$, Sanskriti Aripineni$^{1}$\\
Shijian Deng$^{1}$, Jon E. Froehlich$^{2}$, Yuhang Zhao$^{3}$, Yapeng Tian$^{1}$\\[0.4em]
$^{1}$The University of Texas at Dallas \quad
$^{2}$University of Washington \quad
$^{3}$University of Wisconsin--Madison\\
{\tt\small \{jia.li, wenjie.zhao, fnu.atisri, sanskriti.aripineni, shijian.deng, yapeng.tian\}@utdallas.edu}\\
{\tt\small jonf@cs.washington.edu \quad yuhang.zhao@wisc.edu}
}

\begin{document}
\maketitle

\begin{abstract}
Egocentric gaze estimation is commonly studied using models that process the full video with access to future frames, while real-world applications require strictly causal, online prediction. This discrepancy raises key questions: Does future context inherently provide valuable signals for gaze estimation? If so, how much future look-ahead optimally supervises a causal model during training? To investigate, we propose a controlled framework featuring a future-aware branch that accesses a tunable look-ahead horizon during training but is discarded at inference. This design isolates the impact of future context while keeping the inference architecture fixed and strictly causal. Across EGTEA Gaze+ and Ego4D, we find that future-privileged supervision consistently improves causal gaze prediction, confirming its utility. However, performance gains do not increase monotonically with longer look-ahead, but rather peak within a bounded temporal regime. Specifically, optimal performance corresponds to roughly 1.7--3.3 seconds of future context ($H{\in}[5, 10]$) on EGTEA Gaze+ and 2.7 seconds ($H{=}10$) on Ego4D. Our results demonstrate that lightweight causal models can effectively absorb future-aware signals, providing practical guidance for real-time egocentric gaze modeling.
\end{abstract}

\section{Introduction}
\label{sec:intro}

\begin{figure}[t]
    \centering
    \includegraphics[width=\linewidth]{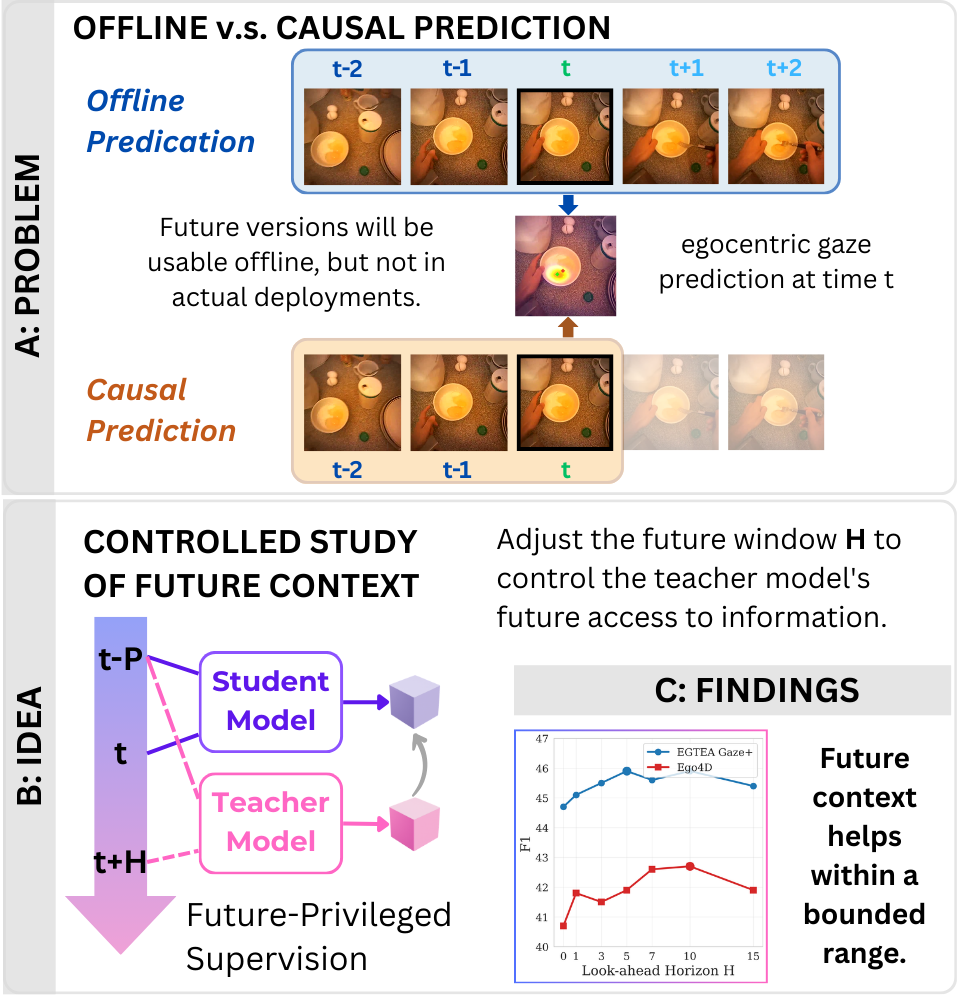}
    \caption{Many existing egocentric gaze models assume offline access to future frames, whereas real-world online systems must predict gaze strictly causally from past and present observations only. To study this mismatch, we introduce a controlled framework in which future context is available only during training through a future-aware branch, while inference remains strictly causal. Across EGTEA Gaze+ and Ego4D, future-privileged supervision consistently improves the causal baseline, although the gains are concentrated within a bounded temporal range rather than growing monotonically with longer look-ahead.}
\label{fig:teaser}
\end{figure}

Egocentric gaze estimation aims to localize the camera wearer’s focus of attention from first-person visual input, and serves as an important component in augmented reality assistance~\cite{lee2024gazepointar, belardinelli2024gaze}, assistive systems, and large-scale attention analysis~\cite{grauman2022ego4d, wang2024gazeprompt}. Over the past decade, performance on egocentric gaze benchmarks has steadily improved, evolving from early methods based on hand-crafted cues, saliency, and short-term temporal modeling~\cite{li2013learning, huang2018predicting} to transformer-based architectures with stronger spatio-temporal reasoning~\cite{lai2023glc}.

However, many existing methods are formulated in offline settings that allow bidirectional temporal access during both training and inference, as shown in Fig.~\ref{fig:teaser}. While some recurrent approaches do support causal inference using only past observations, the offline assumption remains impractical for many real-world applications. In augmented reality and wearable assistance, gaze must be estimated online under strict causality and low-latency constraints, using only past and present observations. The defining difference between offline and online settings is therefore the availability of future frames. This mismatch raises two fundamental questions: \textit{First, does the absence of future context inherently deprive the model of valuable anticipatory cues for gaze prediction? Second, if future information is indeed useful, how much of that temporal context can be successfully distilled into a strictly causal model during training?}


A plausible reason future context matters is that egocentric gaze is not purely reactive, but often anticipatory. During goal-directed activities, people frequently fixate on an object before acting on it~\cite{land1999roles}. More broadly, subsequent frames can reveal evolving hand-object interactions, object state changes, and task progression that make the user’s intent more explicit than the current frame alone. Together, these observations suggest that future context may provide useful supervision for causal gaze prediction. Yet, the field still lacks a clear understanding of how the utility of future context scales with temporal horizon, whether its benefit saturates within a bounded range, and to what extent such information can be transferred to a strictly causal model.

To investigate these questions, we propose a controlled future-privileged training framework for causal egocentric gaze estimation. Our goal is not to redefine gaze modeling itself, but to isolate and quantify how future-privileged supervision affects strictly causal egocentric gaze estimation under controlled conditions. Our framework incorporates a future-aware branch that has access to a tunable look-ahead horizon $H$ during training, but is discarded at inference. The causal and future-aware branches are implemented as two masked forward passes of the same lightweight decoder, differing only in their temporal attention masks. By varying $H$ while keeping the visual encoder and inference-time architecture fixed, we isolate the effect of future context from changes in representation or model capacity.

Empirically, we find that future-privileged supervision improves causal gaze prediction. Across EGTEA Gaze+ and Ego4D, F1 scores improve from the causal baseline ($H{=}0$) of 44.7 to 45.9, and 40.7 to 42.7, respectively. Interestingly, these gains do not scale monotonically with longer look-ahead. Instead, performance peaks within a bounded temporal regime: $H{\in}[5, 10]$ (roughly 1.7--3.3 seconds) for EGTEA Gaze+, and $H{=}10$ (roughly 2.7 seconds) for Ego4D. These findings confirm the utility of future context while suggesting that its range is bounded.

Our main contributions are as follows:
\begin{itemize}[leftmargin=*,topsep=2pt,itemsep=4pt]
\item \textbf{A controlled formulation for studying future context.} We formulate egocentric gaze estimation under a causal online setting and introduce a future-privileged training framework that isolates the impact of future look-ahead while keeping the inference architecture fixed.

\item \textbf{An empirical characterization of the optimal future range.} Across EGTEA Gaze+~\cite{li2021eye} and Ego4D~\cite{grauman2022ego4d}, we demonstrate that future-privileged supervision consistently outperforms the causal baseline, yet the benefits concentrate within a specific, bounded temporal window (1.7-3.3 seconds) rather than increasing indefinitely.

\item \textbf{Practical guidance for online egocentric gaze modeling.} We show that a lightweight causal decoder can effectively absorb future-aware signals during training while preserving strict causality at inference, offering actionable insights for the design of real-time, low-latency egocentric gaze systems.
\end{itemize}
\section{Related Work}
\label{sec:related_work}

\subsection{Egocentric Gaze Estimation}
Egocentric gaze estimation aims to infer a camera wearer’s focus of attention from first-person visual observations, often leveraging cues such as hand-object interactions, object saliency, and temporal context~\cite{fathi2012learning, li2013learning, huang2018predicting}. Recent deep models have significantly improved performance by incorporating richer spatio-temporal reasoning. In particular, transformer-based approaches such as GLC~\cite{lai2023glc} model long-range temporal dependencies and global-local interactions across video clips. 
However, many existing methods operate in an offline setting, relying on bidirectional temporal context during both training and inference. While some recurrent approaches do support causal inference using only past observations, the field still lacks a systematic understanding of whether and how much future observations leveraged during training can enhance strictly causal inference at test time. This gap motivates our controlled investigation of future-privileged supervision for online egocentric gaze estimation.

\subsection{Gaze Anticipation and Anticipatory Gaze Behavior}
A closely related line of work studies \emph{gaze anticipation}, where the goal is to forecast a future gaze location from current or past observations~\cite{lai2024listen}. This task is related to, but fundamentally distinct from, strictly causal egocentric gaze estimation. Gaze anticipation shifts the prediction target to a future time step, whereas our setting predicts the user’s \emph{current} gaze under a causal observation constraint. In our formulation, future frames are not test-time inputs and future gaze is not the prediction target; instead, future observations are used exclusively as a privileged training signal \textit{to investigate whether they can improve} the causal prediction of the current gaze.

The hypothesis that future video frames could inform current gaze prediction stems directly from the inherently anticipatory nature of human vision. Cognitive science studies demonstrate that human gaze is rarely purely reactive. During goal-directed activities, individuals frequently fixate on an object well before actually acting upon it~\cite{land1999roles}. Because the current gaze is intimately tied to upcoming actions, subsequent video frames, which capture these actions as they unfold, often reveal the user's immediate intent much more explicitly than the current frame alone~\cite{belardinelli2022intention, plizzari2024outlook}. While this behavioral link between current gaze and future actions is well-established, the computer vision literature has not yet systematically characterized how the utility of such future context scales temporally when training strictly causal models.

\subsection{Privileged Supervision for Causal Prediction}
Using future information during training to improve causal inference at test time has been explored in several neighboring areas. In action recognition and anticipation, prior work has studied privileged supervision and knowledge distillation, where a causal student is guided by a model or branch with access to future frames during training \cite{zhao2020pkd, zhao2022ppkd}. Related efforts in streaming video modeling have also investigated architectures and memory mechanisms for causal long-video understanding \cite{cheng2024videollama, chen2024videollm, qian2024streaming}. Our work is most closely related to these training-time privileged supervision paradigms, but differs in both task and goal. Rather than introducing a new distillation algorithm, we use future-privileged supervision as a controlled tool to study how the utility of future context changes with look-ahead horizon in egocentric gaze estimation. This perspective is especially relevant for dense, continuous prediction tasks, where the amount of useful future information may not scale in the same way as in clip-level recognition.

\begin{figure*}[t]
    \centering
    \includegraphics[width=\linewidth]{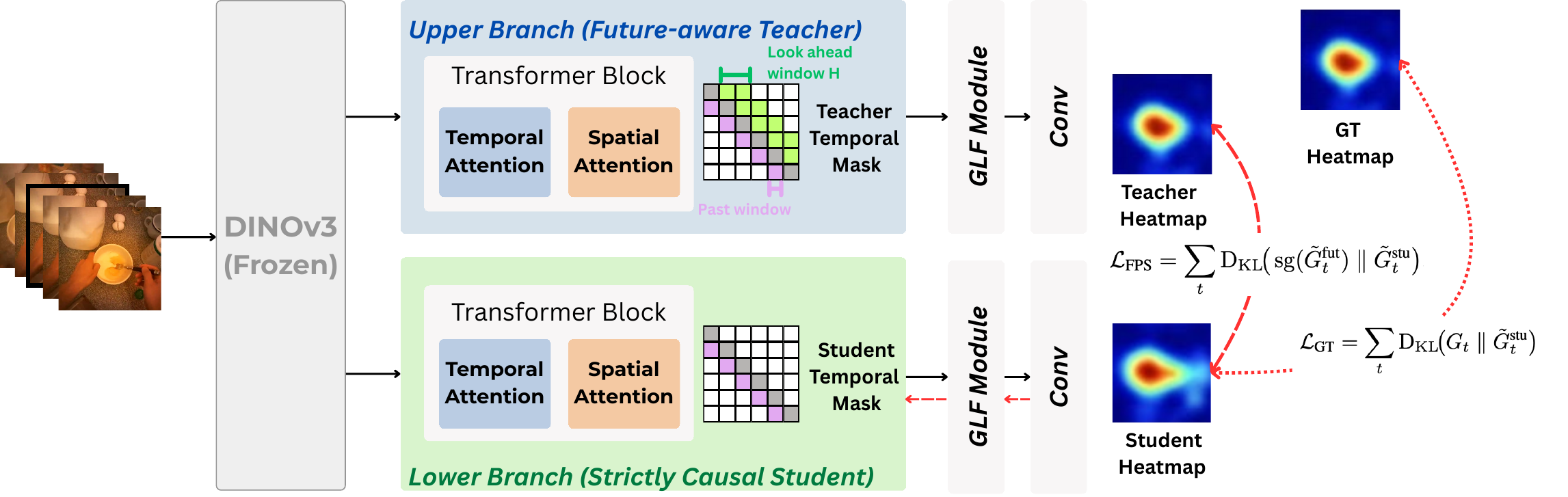}

    \caption{\textbf{Overview of our controlled future-privileged training framework.} A frozen DINOv3 encoder extracts per-frame features, which are fed into two masked forward passes of the same lightweight decoder: a future-aware teacher (upper branch) and a strictly causal student (lower branch).  The two branches share the identical spatio-temporal transformer architecture and parameters, differing exclusively in their temporal attention masks.  The student attends strictly to past and present frames (lower triangular mask), while the teacher is granted a tunable look-ahead window $H$ during training.  Following temporal and spatial attention, features are processed by a GLF module and a convolutional head to yield spatial gaze heatmaps.  During training, the student is optimized using both a standard ground-truth objective ($\mathcal{L}_{\text{GT}}$) and a future-privileged supervision loss ($\mathcal{L}_{\text{FPS}}$), which distills anticipatory knowledge by encouraging the student to match the teacher's stop-gradient predictions. At inference, only the strictly causal student is deployed.}
    \label{fig:model_architecture}
\end{figure*}

\section{ECOGaze: A Controlled Framework for Causal Gaze Prediction}

\subsection{Problem Definition}

\paragraph{Input and target.}
Let $X_{1:T}=(x_1,\dots,x_T)$ be an egocentric RGB video clip with frames $x_t\in\mathbb{R}^{H\times W\times 3}$ and spatial pixel grid $\Omega=\{1,\dots,H\}\times\{1,\dots,W\}$. For each time step $t$, the ground-truth gaze fixation is represented as a spatial probability map $G_t\in\Delta(\Omega)$, where $G_t(i,j)\ge 0$ and $\sum_{(i,j)\in\Omega} G_t(i,j)=1$. The model is tasked with outputting a predicted gaze distribution $\hat G_t\in\Delta(\Omega)$ such that:
\[
\hat G_t = f_\theta(X_{1:t}) \in \Delta(\Omega),\qquad t=1,\dots,T.
\]

\paragraph{Strictly causal constraint.}
In the strictly causal regime, the prediction at time $t$ must depend solely on frames observed up to $t$. Mathematically, this enforces that:
\begin{equation}
\frac{\partial \hat G_t}{\partial x_s}=0\quad \text{for all } s>t.
\label{eq:causal}
\end{equation}
While offline or bidirectional video transformers are typically trained and evaluated on clips $X_{t-L:t+H}$ with a future look-ahead $H\ge0$, which generally inflates performance on standard benchmarks~\cite{bertasius2021space, arnab2021vivit}, this explicitly violates Eq.~\eqref{eq:causal}. Our study rigorously enforces this strictly causal constraint at inference time, reflecting the fundamental latency requirements of interactive AR and real-time assistive applications.

\subsection{Framework Overview}

To systematically investigate how future context affects strictly causal gaze prediction, we propose a controlled architecture, illustrated in Fig.~\ref{fig:model_architecture}. The framework is specifically designed to isolate the effect of future context: we keep the visual representation and inference-time capacity strictly fixed, varying only the temporal look-ahead horizon available during the training phase.

\paragraph{Scene encoder.}
Egocentric gaze is strongly dictated by scene semantics. In everyday interactions, individuals naturally fixate on hands, tools, and manipulated objects~\cite{li2013learning, huang2017jointattention, lai2023glc}. To ensure our gaze predictions are grounded in robust semantic understanding, we adopt a frozen DINOv3 Vision Transformer~\cite{siméoni2025dinov3} as the foundational scene encoder. For each frame, it outputs a grid of patch-level spatial features. Crucially, we keep this encoder \emph{frozen} across all experiments. This intentional design choice eliminates representation-level confounds, ensuring that any performance variations across different temporal settings directly reflect the impact of the future-privileged supervision rather than feature drift.

\paragraph{Spatio-temporal head.}
Because the frozen DINOv3 processes frames independently, it lacks the temporal reasoning necessary to interpret evolving actions and short-term eye-hand coordination. To model these dynamics, we append a lightweight spatio-temporal transformer head. 
Let $\mathbf{X}^{(0)}$ denote the sequence of per-frame patch embeddings from the encoder. We process $\mathbf{X}^{(0)}$ using $L$ layers of divided space-time attention~\cite{bertasius2021space}. Within each layer, temporal attention first aggregates short-range evidence across frames, followed by spatial attention to model interactions among intra-frame patch tokens. 
We intentionally keep this temporal head lightweight to ensure our study focuses purely on the utility of future context, rather than simply scaling up decoder capacity.

\paragraph{Future-privileged supervision.}
This is the core mechanism of our controlled study. To leverage future context during training without violating causality at inference, we implement a dual-branch forward pass using the \emph{exact same} spatio-temporal decoder, differing only in their temporal attention masks. 

Specifically, for a given look-ahead horizon $H$, the framework executes two passes: 1) A \textbf{future-aware teacher branch}, which uses an expanded temporal mask that allows each token to attend to past frames, the current frame, and $H$ future frames. 2) A \textbf{strictly causal student branch}, which uses a strict lower-triangular temporal mask ($H=0$), restricting attention to past and present frames.

Because both branches share identical parameters, this masked asymmetry allows us to vary the amount of future context ($H$) in a controlled manner. 
We deliberately adopt this isomorphic design rather than a separately trained, stronger teacher: a more powerful teacher would confound the analysis, as any observed gain could reflect the teacher's superior capacity rather than the utility of future temporal access itself. By keeping both branches parameter-shared and architecturally identical, we ensure that the sole variable is the temporal attention mask. The causal student serves as the final model deployed at inference time, with the future-aware mask completely discarded.

\paragraph{Global-local focusing and prediction.}
Following prior work~\cite{lai2023glc}, we employ a lightweight global-local focusing (GLF) module to refine the features. For each frame, a query vector formed by a learnable prior and the frame's global token computes cosine similarities with all patch features from the final transformer layer. This produces a residual gate $\boldsymbol{\alpha}_t$ that highlights gaze-relevant regions:

\[
\mathbf{X}^{\text{focus}}_{t,n} = \mathbf{X}^{(L)}_{t,n} + \alpha_{t,n}\,\mathbf{X}^{(L)}_{t,n}.
\]
Finally, a compact convolutional prediction head reshapes and upsamples the focused tokens, applying a $1 \times 1$ convolution and temperature-scaled softmax to yield the final per-frame spatial probability map $\hat G_t \in \Delta(\Omega)$. 

\subsection{Training Objective}
During training, the strictly causal student is optimized using both ground-truth labels and a distillation signal from the future-aware teacher. Because both branches are simply differently-masked forward passes of the same network, this privileged supervision requires no auxiliary teacher models.

Let $\mathrm{softmax}_\tau(z)=\mathrm{softmax}(z/\tau)$ denote temperature scaling, with $\tau{=}2$ in all experiments. We obtain smoothed distributions for the student and the future-aware teacher:
$\tilde G^{\text{stu}}_{t}=\mathrm{softmax}_\tau(\hat G^{\text{stu}}_{t})$ and 
$\tilde G^{\text{fut}}_{t}=\mathrm{softmax}_\tau(\hat G^{\text{fut}}_{t})$.

The overall objective minimizes the combination of a standard ground-truth loss and a future-privileged supervision (FPS) loss:
\[
\mathcal{L}
= \alpha\,\mathcal{L}_{\text{GT}}
+ \beta\,\mathcal{L}_{\text{FPS}},
\]
where the components are defined via KL divergence over the spatial grid:
\[
\mathcal{L}_{\text{GT}}
=\sum_{t}\mathrm{D_{KL}}\!\big(G_t\;\Vert\;\tilde G^{\text{stu}}_{t}\big),
\]
\[
\mathcal{L}_{\text{FPS}}
=\sum_{t}\mathrm{D_{KL}}\!\big(\operatorname{sg}(\tilde G^{\text{fut}}_{t})
\;\Vert\;\tilde G^{\text{stu}}_{t}\big).
\]
Crucially, the stop-gradient operator $\operatorname{sg}(\cdot)$ is applied to the future-aware branch. This ensures that the shared decoder weights are updated primarily to improve the \emph{student's} causal prediction, while being guided to anticipate the future-aware distribution. We set $\alpha=1$ and $\beta=1$ empirically across all experiments.
\section{Experiments}
\label{sec:exp}

\begin{figure*}[t]
    \centering
    \includegraphics[width=\textwidth]{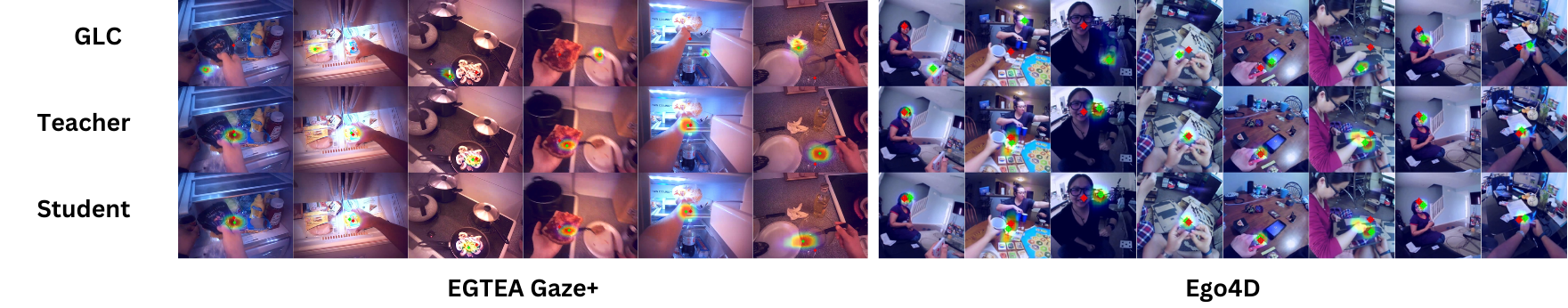}
    \caption{Qualitative comparison of gaze predictions on EGTEA Gaze+ and Ego4D under the strictly causal setting. From top to bottom: the causal variant of GLC, the training-time future-aware teacher ($H>0$), and our strictly causal student (ECOGaze, $H=0$). In many cases, the strictly causal student successfully internalizes future-guided attention patterns, producing more accurate focal maps than the causal GLC baseline.}
    \label{fig:teacher_student_vis}
\end{figure*}

\subsection{Experimental Setup}
\paragraph{Dataset.}
We conduct experiments on two egocentric gaze benchmarks. \textbf{EGTEA Gaze+}~\cite{li2021eye} contains 28 hours of egocentric cooking activities recorded at 24 FPS. Following the official protocol, we use 8,299 clips for training and 2,022 clips for testing. \textbf{Ego4D (gaze subset)}~\cite{grauman2022ego4d} is a large-scale egocentric benchmark, from which we use the gaze-annotated subset with normalized 2D gaze coordinates. Following prior work~\cite{lai2023glc}, gaze annotations are converted into $64\times64$ heatmaps for supervision and evaluation.

\paragraph{Evaluation Metrics.}
Following conventions~\cite{lai2023glc, li2021eye}, we report \textbf{Adaptive F1 Score}, \textbf{Precision}, and \textbf{Recall}. These metrics measure the overlap between predicted heatmaps and the ground truth. Specifically, predicted heatmaps are thresholded over $\tau \in [0, 0.02]$. All metrics are computed exclusively on frames with valid gaze annotations.

\paragraph{Implementation Details.}
All models are implemented in PyTorch~\cite{paszke2019pytorch} and trained on a single NVIDIA A5000 GPU. We adopt the DINOv3-ViT-S/16 encoder pretrained on the LVD-1689M dataset from Meta AI. The backbone remains frozen and outputs 384-dimensional features. Models are trained for 25 epochs on EGTEA Gaze+ and 15 epochs on Ego4D with a batch size of 16. We use the AdamW optimizer with a learning rate of $1\times10^{-4}$, a weight decay of 0.05, and a cosine learning-rate schedule applied after five warm-up epochs. Input frames are resized to $224\times224$ with standard ImageNet normalization. Across all scaling experiments, we keep the encoder, spatio-temporal head, prediction head, and optimization setup fixed, varying only the look-ahead horizon $H$ used by the future-aware teacher during training.

\subsection{Controlled Study Protocol}
Our study is designed to rigidly isolate the effect of future context from other sources of variation. By freezing the encoder and fixing the causal student’s inference architecture, we eliminate representational and capacity-related confounds. Any performance variance can thus be attributed to the amount of future-privileged supervision.

We vary the tunable look-ahead horizon as follows:
\[
H \in \{0, 1, 3, 5, 7, 10, 15\}.
\]
On EGTEA, videos are sampled at 24 FPS, and neighboring temporal tokens are separated by a stride of 8 frames. On Ego4D, videos are sampled at 30 FPS with the same temporal stride. Under these settings, one future step ($H=1$) corresponds to approximately 0.33 seconds on EGTEA and 0.27 seconds on Ego4D.

This protocol aims to answer our central research questions: \textit{Does a strictly causal gaze predictor benefit from future-privileged supervision? And if so, do the gains continue to grow with longer look-ahead, or do they saturate within a bounded temporal window?}

\subsection{How Much Future Context Helps}
We first analyze how the amount of training-time future context affects the causal student's inference performance, using F1 as the primary metric.

\paragraph{Performance trends.}
Table~\ref{tab:scaling_dual} summarizes the effect of varying the look-ahead horizon $H$. On \textbf{EGTEA Gaze+}, future-privileged supervision consistently improves performance over the causal baseline ($H{=}0$) across all nonzero horizons. The F1 score increases from 44.7 at $H{=}0$ to 45.5 at $H{=}3$, peaking at 45.9 for both $H{=}5$ and $H{=}10$. Between these two peak settings, $H{=}5$ yields the highest precision (36.7), while $H{=}10$ gives the highest recall (63.6). Notably, at $H{=}15$, performance slightly drops to 45.4 F1. 
On \textbf{Ego4D}, we observe a similar, albeit more gradual, positive trend. Relative to the baseline (40.7 F1), most nonzero horizons enhance performance, with the absolute best F1 (42.7) and recall (57.6) achieved at $H{=}10$. Similar to EGTEA, the gain degrades at $H{=}15$, where F1 drops to 41.9. Together, both datasets reveal a consistent pattern: \emph{future context definitively aids strictly causal prediction, but more look-ahead is not always better. The benefits peak within a specific bounded range and diminish thereafter.}

\paragraph{Temporal interpretation.}
These quantitative trends become highly interpretable when $H$ is mapped to real-world time. With our frame stride, the optimal F1 on EGTEA ($H{\in}[5, 10]$) corresponds to roughly 1.67--3.33 seconds of future context, while the Ego4D peak ($H{=}10$) corresponds to roughly 2.67 seconds. Across both diverse datasets, the most useful privileged supervision is extracted from a future window of approximately 2 to 3 seconds.

This finding is particularly intriguing in light of cognitive science literature, which suggests that human eye-hand coordination operates at sub-second timescales (roughly 500--1000\,ms)~\cite{land1999roles}. Our results indicate that the value of future supervision extends beyond immediate anticipatory latency; it successfully captures longer-range task unfolding and object state changes (e.g., reaching for and successfully grasping a tool) that typically complete within a 2-3 second window. However, pushing the horizon too far ($H=15$, $\approx$ 4-5 seconds) introduces noise from subsequent, unrelated actions, thereby degrading the distillation signal.

\begin{table}[t]
\centering
\footnotesize
\begin{tabular}{l|ccc|ccc}
\toprule
 & \multicolumn{3}{c|}{\textbf{EGTEA Gaze+}} & \multicolumn{3}{c}{\textbf{Ego4D}} \\
\cmidrule(lr){2-4} \cmidrule(lr){5-7}
\textbf{H} & \textbf{F1} & \textbf{Rec.} & \textbf{Prec.} & \textbf{F1} & \textbf{Rec.} & \textbf{Prec.} \\
\midrule
H = 0 (Baseline) & 44.7 & 60.3 & 35.5 & 40.7 & 56.3 & 31.9 \\
H = 1  & 45.1 & 59.2 & 36.4 & 41.8 & 56.1 & 33.4 \\
H = 3  & 45.5 & \textbf{61.5} & 36.1 & 41.5 & 56.8 & 32.7 \\
H = 5  & \textbf{45.9} & 61.1 & \textbf{36.7} & 41.9 & 55.7 & 33.6 \\
H = 7  & 45.6 & 60.3 & 36.6 & 42.6 & 56.9 & \textbf{34.0} \\
H = 10 & \textbf{45.9} & \textbf{63.6} & 35.9 & \textbf{42.7} & \textbf{57.6} & \textbf{34.0} \\
H = 15 & 45.4 & 61.5 & 36.1 & 41.9 & 56.6 & 33.2 \\
\bottomrule
\end{tabular}
\caption{Effect of varying the look-ahead horizon $H$ on strictly causal gaze prediction. Future-privileged supervision improves performance on both benchmarks, peaking around $H{\in}[5, 10]$ before degrading at extreme horizons ($H{=}15$).}
\label{tab:scaling_dual}
\end{table}

\subsection{Competitive Performance Under Causality}
To contextualize the practical value of our controlled framework, we designate our best-performing causal student as \textbf{ECOGaze} and compare it against representative prior methods. As shown in Table~\ref{tab:gaze_comparison}, the comparison spans classical saliency-based approaches, earlier task-specific models, and recent transformer-based architectures. For fair comparison under strict causality, we include a causal variant of the state-of-the-art GLC model~\cite{lai2023glc}, retrained to use only past and present frames.

For ECOGaze, we deploy the $H{=}5$ variant for EGTEA Gaze+ (as it attains the peak F1 with higher precision and less future reliance) and the $H{=}10$ variant for Ego4D. Under these settings, ECOGaze performs exceptionally well. On \textbf{EGTEA Gaze+}, it achieves a 45.9 F1 score, outperforming all reported baselines by a significant margin. On \textbf{Ego4D}, it likewise achieves the highest F1 (42.7) and recall (57.6) while maintaining competitive precision. These results confirm that our framework is not just an analytical tool for studying future context, but a robust methodology for deriving highly competitive causal predictors.

\newcolumntype{C}{>{\centering\arraybackslash}X}
\begin{table}[t]
\centering
\footnotesize
\begin{tabularx}{\columnwidth}{l|CCC|CCC}
\toprule
 &
\multicolumn{3}{c|}{\textbf{EGTEA Gaze+}} &
\multicolumn{3}{c}{\textbf{Ego4D}} \\
\cmidrule(lr){2-4} \cmidrule(lr){5-7}
\textbf{Method} & \textbf{F1} & \textbf{Rec.} & \textbf{Prec.}
& \textbf{F1} & \textbf{Rec.} & \textbf{Prec.} \\
\midrule
Center Prior & 10.7 & 32.0 & 6.4 & 14.9 & 21.9 & 11.3 \\
GBVS \cite{harel2006graph} & 15.7 & 45.1 & 9.5 & 18.0 & 47.2 & 11.1 \\
EgoGaze \cite{li2013learning} & 16.3 & 16.3 & 16.3 & -- & -- & -- \\
Gaze MLE\textsuperscript{\dag} \cite{li2021eye} & 26.6 & 35.7 & 21.3 & -- & -- & -- \\
Joint Learning\textsuperscript{\dag} \cite{li2021eye} & 34.0 & 42.7 & 28.3 & -- & -- & -- \\
I3D-R50\textsuperscript{\dag} \cite{feichtenhofer2019slowfast} & 40.9 & 57.2 & 31.8 & -- & -- & -- \\
Attention Transition \cite{huang2018predicting} & 37.2 & 51.9 & 29.0 & 36.4 & 47.5 & 29.5 \\
GLC (Causal) \cite{lai2023glc} & 41.6 & 57.9 & 32.4 & 41.2 & 56.1 & 32.5 \\
\rowcolor{rowgray}
ECOGaze (Ours) & \textbf{45.9} & \textbf{61.1} & \textbf{36.7} & \textbf{42.7} & \textbf{57.6} & \textbf{34.0} \\
\bottomrule
\end{tabularx}
\caption{Comparison with prior methods for egocentric gaze prediction. The baseline models marked with $\dag$ use gaze supervision alongside action labels. Our causal student (ECOGaze) is competitive with prior approaches and performs favorably on both benchmarks.}
\label{tab:gaze_comparison}
\end{table}

\subsection{Accuracy-Efficiency Trade-off}
\begin{figure}[t]
\centering
\includegraphics[width=0.58\linewidth]{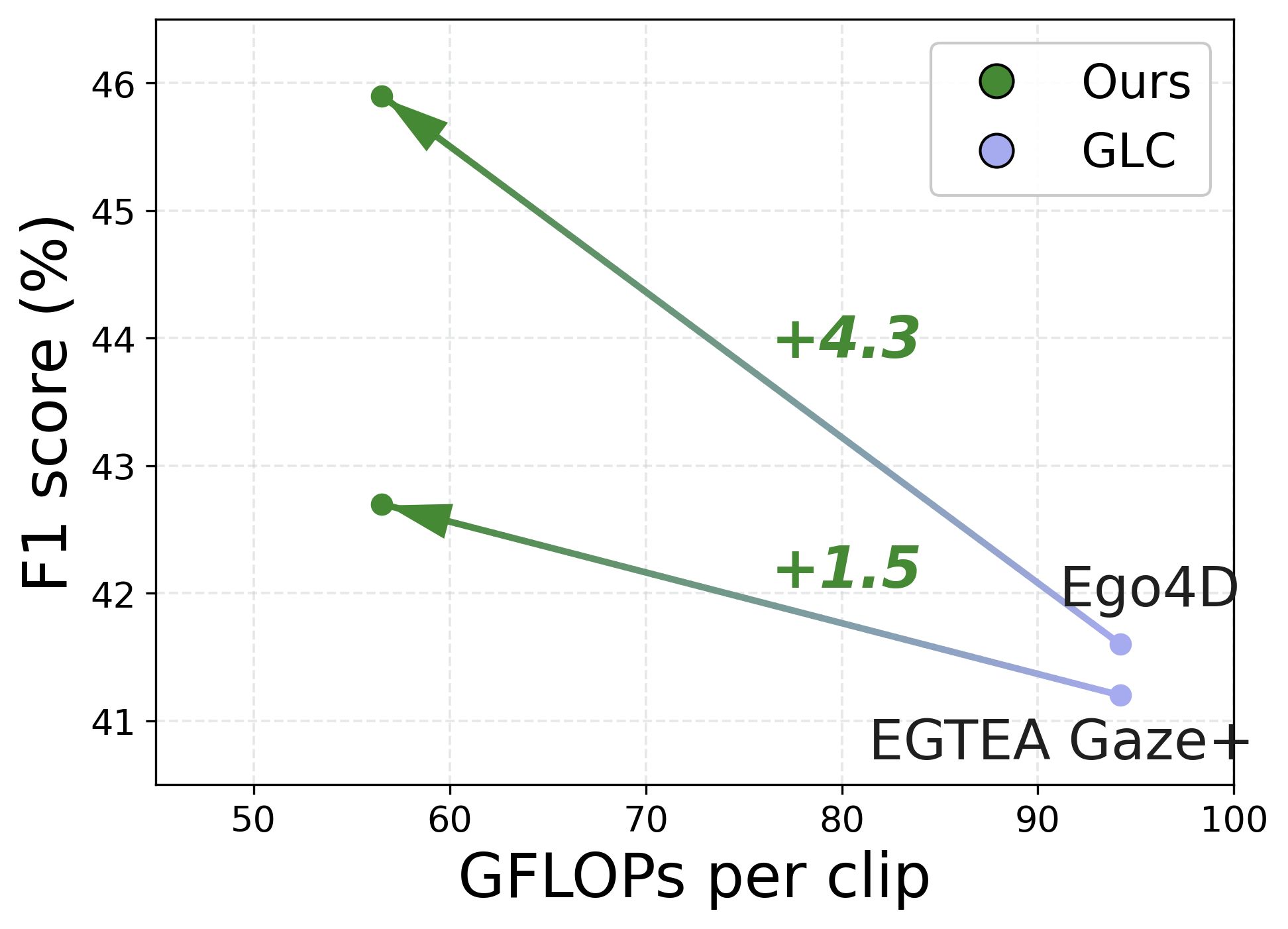}
\caption{
Accuracy-efficiency trade-off on EGTEA Gaze+ and Ego4D. ECOGaze attains higher F1 scores while requiring fewer GFLOPs per clip compared to the GLC baseline.}
\label{fig:efficiency1}
\end{figure}

Finally, we evaluate ECOGaze in terms of computational efficiency, a critical requirement for real-time systems. Fig.~\ref{fig:efficiency1} plots the F1 score against GFLOPs per clip. On both benchmarks, ECOGaze comfortably dominates the Pareto frontier, attaining substantially higher F1 scores while requiring fewer GFLOPs than the causal GLC baseline.

Table~\ref{tab:efficiency2} further details model size and inference throughput. Thanks to our intentional lightweight decoder design, ECOGaze contains only 14.2M trainable parameters, compared to GLC's 70.2M, a nearly $5\times$ reduction in capacity. When deployed on a single NVIDIA A5000 GPU, ECOGaze runs at roughly 60 FPS, providing a $2\times$ speedup over GLC ($\approx$ 30 FPS). Although our framework was primarily architected for controlled empirical analysis, these system-level metrics prove that it simultaneously offers a highly compact, low-latency solution ideal for real-time egocentric gaze applications.

\begin{table}[t]
\centering
\footnotesize
\setlength{\tabcolsep}{6pt}
\renewcommand{\arraystretch}{1.1}
\begin{tabular}{l|cc|cc}
\toprule
& \multicolumn{2}{c|}{\textbf{EGTEA Gaze+}} & \multicolumn{2}{c}{\textbf{Ego4D}} \\
\cmidrule(lr){2-3} \cmidrule(lr){4-5}
\textbf{Model} & \textbf{Para. (M)} & \textbf{FPS} $\uparrow$ & \textbf{Para. (M)} & \textbf{FPS} $\uparrow$ \\
\midrule
GLC (Causal) & 70.18 & 30.28 & 70.18 & 29.27 \\
ECOGaze (Ours) & \textbf{14.19} & \textbf{59.01} & \textbf{14.19} & \textbf{59.73} \\
\bottomrule
\end{tabular}
\caption{Comparison of model size and inference throughput. ECOGaze is substantially smaller ($5\times$ fewer parameters) and faster ($2\times$ throughput) than GLC while remaining accurate.}
\label{tab:efficiency2}
\end{table}

\section{Ablation Studies}
\label{sec:ablation}

\subsection{Component-wise Analysis}
We first analyze how the four core components of our controlled framework contribute to the final performance on EGTEA Gaze+: Spatial Attention (SAttn), Temporal Attention (TAttn), Global-Local Focusing (GLF), and Future-Privileged Supervision (FPS). 

As shown in Table~\ref{tab:egtea_ablation}, the baseline utilizing only static features achieves a modest 39.1 F1 score. Adding spatial attention significantly improves F1 to 42.3, indicating that modeling spatial interactions among patch tokens is crucial even when using strong, frozen visual features. Integrating temporal attention further raises F1 to 44.4, demonstrating that short-range temporal cues (e.g., motion and action dynamics) provide vital information beyond static appearance. Adding the lightweight global-local focusing module yields another gain, bringing F1 to 44.7, which confirms the value of scene-level guidance. Finally, incorporating future-privileged supervision boosts F1 to the peak of 45.9, validating our core hypothesis that future-aware training signals distill anticipatory knowledge into the causal student.

Interestingly, the static baseline attains the highest recall (63.9) but suffers from substantially lower precision (28.2), indicating scattered, less accurate localization. As structured reasoning is added to the causal decoder, precision improves steadily while recall remains competitive, ultimately producing a stronger and more balanced F1 score.

\begin{table}[t]
\centering
\small
\begin{tabular}{ccccccc}
\toprule
\textbf{SAttn} & \textbf{TAttn} & \textbf{GLF} & \textbf{FPS} & \textbf{F1} & \textbf{Rec.} & \textbf{Prec.} \\
\midrule
\xmark & \xmark & \xmark & \xmark & 39.1 & \textbf{63.9} & 28.2 \\
\cmark & \xmark & \xmark & \xmark & 42.3 & 58.8 & 33.0 \\
\cmark & \cmark & \xmark & \xmark & 44.4 & 58.7 & 35.8 \\
\cmark & \cmark & \cmark & \xmark & 44.7 & 60.3 & 35.5 \\
\cmark & \cmark & \cmark & \cmark & \textbf{45.9} & 61.1 & \textbf{36.7} \\
\bottomrule
\end{tabular}
\caption{Component-wise ablation on EGTEA Gaze+. Spatial attention (SAttn), temporal attention (TAttn), global-local focusing (GLF), and future-privileged supervision (FPS) cumulatively improve performance, with the full framework achieving the strongest F1 and precision.}
\label{tab:egtea_ablation}
\end{table}

\subsection{Design Choices Within the Framework}
To ensure our causal student is both accurate and computationally suited for real-time deployment, we examined two architectural design choices: the implementation of global-local focusing and the form of spatio-temporal aggregation. These comparisons justify the lightweight backbone utilized throughout our controlled study.

\begin{figure}[t]
    \centering
    \includegraphics[width=\linewidth]{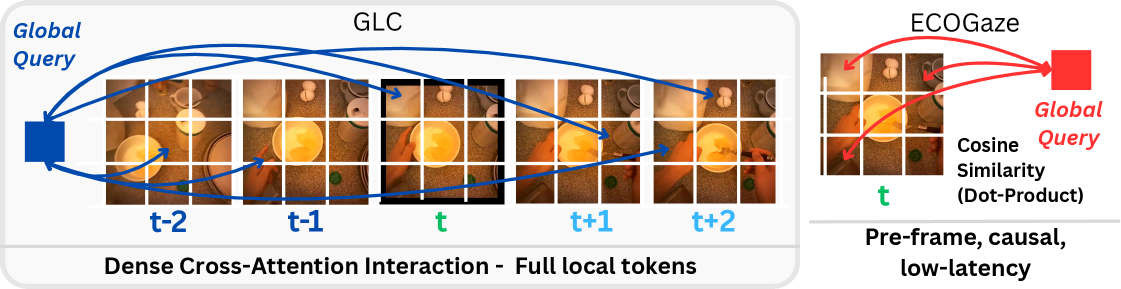}
    \caption{Comparison of global-local focusing designs. GLC (left) uses dense cross-attention between a global query and all local tokens across time. ECOGaze (right) applies a lightweight per-frame causal refinement based on dot-product similarity within the current frame, preserving global guidance while drastically reducing computational cost.}
    \label{fig:gquery}
\end{figure}

\paragraph{Global-local focusing.}
We compare two implementations of the GLF mechanism. As illustrated in Fig.~\ref{fig:gquery}, prior work like GLC~\cite{lai2023glc} employs a dense cross-attention interaction, where a per-frame query attends to all spatial patches across time through computationally heavy learnable projections. In contrast, ECOGaze uses a simpler dot-product formulation: a normalized global query directly computes cosine similarity with patch features strictly within the current frame to obtain aggregation weights. This lightweight design preserves essential global guidance while deliberately avoiding dense cross-frame interactions.

As detailed in Table~\ref{tab:global_query_efficiency}, the dot-product formulation not only improves accuracy on EGTEA Gaze+ (increasing F1 from 45.2 to 45.9 and precision from 36.0 to 36.7) but is also vastly more efficient. It reduces parameter count from 741.1K to merely 149.0K, and slashes FLOPs per clip from 465.4M to a negligible 1.8M, while also slightly lowering peak memory usage. These results unequivocally support the dot-product variant as the optimal GLF design.

\begin{table}[t]
\centering
\small
\begin{tabular}{lcc}
\toprule
\textbf{Metric} & \textbf{Cross-Attn (GLC)} & \textbf{Dot-Product (Ours)} \\
\midrule
F1-score & 45.2 & \textbf{45.9} \\
Precision & 36.0 & \textbf{36.7} \\
Recall & 60.5 & \textbf{61.1} \\
\midrule
Params (K) & 741.1 & \textbf{149.0} \\
FLOPs / clip (M) & 465.4 & \textbf{1.8} \\
FLOPs / frame (M) & 58.2 & \textbf{0.2} \\
Peak Memory (MB) & 23.5 & \textbf{21.8} \\
\bottomrule
\end{tabular}
\caption{Efficiency and accuracy comparison of GLF designs on EGTEA Gaze+. Our dot-product formulation yields slightly stronger accuracy with drastically lower computational overhead.}
\label{tab:global_query_efficiency}
\end{table}

\paragraph{Spatio-temporal aggregation.}
We further compare three distinct designs for spatio-temporal aggregation under strictly causal constraints: a GRU-based temporal head followed by spatial refinement, full Joint Spatio-Temporal attention, and our Divided Temporal-to-Spatial (T+S) attention. Results are presented in Table~\ref{tab:method_comparison}.

The GRU+Spatial variant performs worst overall, achieving a 42.4 F1 score and 32.2 precision. Although it attains the highest recall (62.1), this suggests a tendency to over-predict, resulting in imprecise localization. Joint Spatio-Temporal attention improves accuracy to 45.2 F1 and 35.7 precision, but suffers from an excessively high peak memory footprint (823.27 MB). Our Divided T+S design strikes the best balance, achieving the strongest overall accuracy (45.9 F1 and 36.7 precision) while keeping peak memory low at 292.36 MB.


\paragraph{Additional Diagnostic Analyses.} Due to space constraints, we provide further diagnostic studies in the \textbf{Appendix}. These include a strict \emph{backbone-fair comparison} (isolating the impact of the frozen DINOv3 encoder from our distillation framework) and an evaluation of \emph{robustness to hand visibility} (confirming our model learns deep anticipatory intent rather than shallow hand-tracking biases).

\begin{table}[t]
\centering
\small
\renewcommand{\arraystretch}{1.1}
\resizebox{\columnwidth}{!}{
\begin{tabular}{lccc}
\toprule
\textbf{Metric} & \textbf{GRU+Spatial} & \textbf{Joint ST} & \textbf{Divided T+S (Ours)} \\
\midrule
F1-score & 42.4 & 45.2 & \textbf{45.9} \\
Precision & 32.2 & 35.7 & \textbf{36.7} \\
Recall & \textbf{62.1} & 61.5 & 61.1 \\
\midrule
Params (M) & 10.65 & 5.32 & 10.65 \\
FLOPs / clip (G) & 9.05 & 11.17 & 17.41 \\
FLOPs / frame (G) & 1.13 & 1.40 & 2.18 \\
Peak Memory (MB) & 295.74 & 823.27 & \textbf{292.36} \\
\bottomrule
\end{tabular}}
\caption{Comparison of spatio-temporal aggregation designs on EGTEA Gaze+. Our Divided T+S attention achieves the strongest overall accuracy while remaining highly memory efficient compared to joint attention.}
\label{tab:method_comparison}
\end{table}

\subsection{Failure Case Analysis}
\begin{figure}
    \centering
    \includegraphics[width=\linewidth]{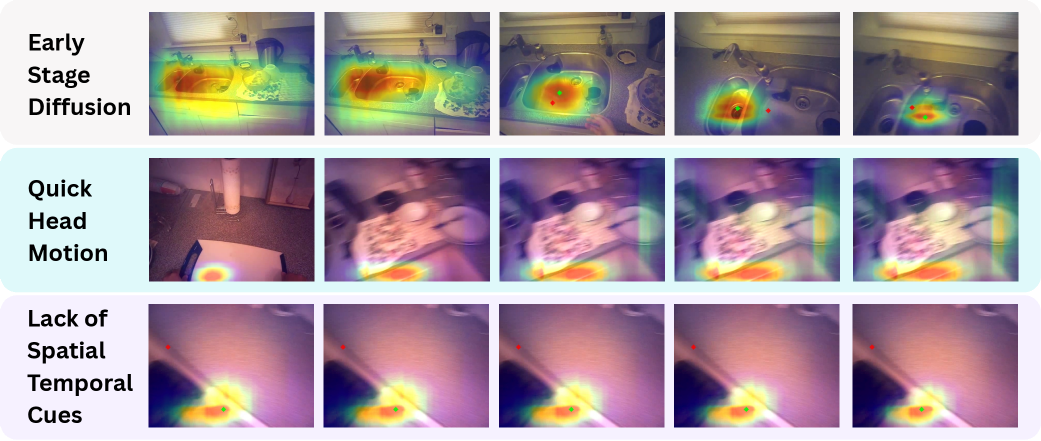}
    \caption{\textbf{Typical failure modes of ECOGaze under the strictly causal online setting.} Lacking future context at inference time, the model occasionally struggles with: (1) \textit{early-stage gaze diffusion} during visual search before a clear fixation occurs; (2) \textit{motion blur} from rapid head movements that corrupt visual features; and (3) \textit{target ambiguity} in cluttered environments, where past and present observations are insufficient to resolve the user's intent.}
    \label{fig:failure_cases}
\end{figure}

While ECOGaze demonstrates robust causal prediction, we identify certain challenging scenarios where the absence of future context at inference time leads to performance degradation. As illustrated in Fig.~\ref{fig:failure_cases}, typical failure modes include: (1) \textit{Early-stage gaze diffusion}, where the user is searching for a target but has not yet fixated, leading to overly broad predictions; (2) \textit{Motion blur} caused by rapid head rotations, which severely degrades the frozen DINOv3 features; and (3) \textit{Target ambiguity} in cluttered scenes with multiple potential objects, where short-term causal history is insufficient to deduce the exact intent. Acknowledging these limitations paves the way for future work integrating longer-term memory or multi-modal cues.

\section{Conclusion and Limitations}

We presented ECOGaze, a controlled future-privileged training framework for strictly causal egocentric gaze estimation. Across EGTEA Gaze+ and Ego4D, future-privileged supervision consistently improves the causal baseline, with the strongest gains appearing within a bounded future range of approximately 1.7 to 3.3 seconds.

\noindent
\textbf{Limitations and Future Work.}
Our study is intended as a controlled analysis of future-privileged supervision rather than a complete account of gaze modeling. The optimal look-ahead range may vary across backbones, model scales, and application domains; in particular, gaze behavior is inherently task-dependent, and the bounded temporal regime identified here may shift across activity types beyond those represented in EGTEA Gaze+ and Ego4D. Extending this analysis from current gaze prediction to \emph{future gaze anticipation}, and evaluating across more diverse activity domains remain important directions for future work.

\clearpage

\section*{Acknowledgments}

This work was supported by the National Eye Institute of the National Institutes of Health under Award Number R01EY037100. The content is solely the responsibility of the authors and does not necessarily represent the official views of the National Institutes of Health.

{
    \small
    \bibliographystyle{ieeenat_fullname}
    \bibliography{main}
}

\clearpage

\section{Appendix}

\subsection{What Does the Causal Model Learn from the Future?}
\label{sec:diagnostic_analysis}

To further validate our empirical findings and understand exactly \emph{why} future-privileged supervision improves causal prediction, we conduct two deeper diagnostic analyses on EGTEA Gaze+.

\paragraph{Isolating the Impact of the Foundation Model.}
A critical necessity in our controlled study is ensuring that the observed performance gains are genuinely driven by the future-aware distillation, rather than merely stemming from the powerful frozen DINOv3 encoder. To strictly isolate this, we re-implemented the causal GLC baseline~\cite{lai2023glc} equipped with the \emph{exact same} frozen DINOv3 backbone. 

As shown in Table~\ref{tab:dinov3_fairness}, the DINOv3-equipped GLC achieves only 33.1 F1, lagging significantly behind our distilled causal student (45.9 F1), despite requiring over $13\times$ more FLOPs. This drastic performance gap confirms a key finding: deploying a strong foundation model is, by itself, insufficient for solving online causal gaze prediction. The true driver of the performance boost is the successful distillation of future context coupled with an efficient decoder, validating the efficacy of our controlled future-privileged framework.

\begin{table}[h]
\centering
\small
\begin{tabular}{lcccc}
\toprule
\textbf{Method} & \textbf{Backbone} & \textbf{F1} $\uparrow$ & \textbf{FLOPs (G)} $\downarrow$ & \textbf{FPS} $\uparrow$ \\
\midrule
GLC (Causal) & DINOv3 & 33.1 & 226.9 & 40.9 \\
\rowcolor{rowgray} 
ECOGaze & DINOv3 & \textbf{45.9} & \textbf{17.4} & \textbf{59.0} \\
\bottomrule
\end{tabular}
\caption{Impact of the foundation model on EGTEA Gaze+. Even when equipped with the exact same frozen DINOv3 encoder, the causal GLC baseline significantly underperforms ECOGaze, confirming that our gains stem fundamentally from future-privileged distillation and our efficient decoder design.}
\label{tab:dinov3_fairness}
\end{table}

\paragraph{Beyond Hand-Tracking: Robustness to Hand Visibility.}
In egocentric videos, gaze is often strongly correlated with hand movements. This raises an important question regarding what the causal student actually learns from the future: \emph{Does future-privileged supervision merely teach the model a shallow "hand-tracking bias," or does it impart deeper anticipatory intent?} 

To answer this, we partitioned the EGTEA Gaze+ test set into "With Hands" and "Without Hands" subsets. As detailed in Table~\ref{tab:hand_occlusion_robustness}, even when hands are entirely absent from the current frame, our causal student maintains a robust F1 score of 44.8 (compared to 39.9 F1 for the GLC baseline) and a high precision of 59.4. This resilience provides a crucial empirical insight: the future context distills broader scene semantics and high-level task progression (e.g., relevant object affordances) into the causal model, enabling robust intent prediction even when explicit hand-object interaction cues are temporarily invisible.

\begin{table}[h]
\centering
\footnotesize
\setlength{\tabcolsep}{4pt}
\begin{tabular}{l|ccc|ccc}
\toprule
\multirow{2}{*}{\textbf{Method}} &
\multicolumn{3}{c|}{\textbf{With Hands}} &
\multicolumn{3}{c}{\textbf{Without Hands}} \\
\cmidrule(lr){2-4} \cmidrule(lr){5-7}
& \textbf{F1} & \textbf{Prec.} & \textbf{Rec.} &
\textbf{F1} & \textbf{Prec.} & \textbf{Rec.} \\
\midrule
Attention Transition \cite{huang2018predicting} & 26.6 & 18.4 & 47.5 & 19.6 & 15.7 & 26.0 \\
GLC (Causal) \cite{lai2023glc} & 43.9 & 34.7 & 59.6 & 39.9 & 31.0 & 56.2 \\
\rowcolor{rowgray} 
ECOGaze (Ours) & \textbf{45.9} & \textbf{36.8} & \textbf{60.1} & \textbf{44.8} & \textbf{59.4} & \textbf{35.9} \\
\bottomrule
\end{tabular}
\caption{Robustness to hand visibility on EGTEA Gaze+. ECOGaze maintains a substantial performance lead over the baselines even in frames where hands are entirely absent, demonstrating that it learns deeper anticipatory intent beyond shallow hand-tracking biases.}
\label{tab:hand_occlusion_robustness}
\end{table}
\end{document}